\documentclass[runningheads]{llncs}
\usepackage{graphicx}
\usepackage{tikz}
\usepackage{comment}
\usepackage{amsmath,amssymb} 
\usepackage{color}

\usepackage{epsfig}
\usepackage{multirow}
\usepackage{makecell}

\graphicspath{{./figures/}}

\begin{document}
\pagestyle{headings}
\mainmatter

\title{Adaptive Object Detection with Dual Multi-Label Prediction }

\titlerunning{Adaptive Object Detection }
%
\author{Zhen Zhao\inst{1} \and
Yuhong Guo\inst{1,2} \and
Haifeng Shen\inst{1} \and
Jieping Ye\inst{1} 
}
\authorrunning{Z. Zhao et al.}
%
\institute{
$^1$\; DiDi Chuxing \qquad\qquad $^2$\; Carleton University\\ 
}
\maketitle

\begin{abstract}

In this paper, we propose a novel end-to-end unsupervised deep domain adaptation model 
for adaptive object detection
by exploiting multi-label object recognition as a dual auxiliary task.
The model exploits multi-label prediction to reveal the object category information in each image
and then uses the prediction results to perform
conditional adversarial global feature alignment,
such that the multimodal structure of
image features can be tackled to 
bridge the domain divergence at the global feature level
while preserving the discriminability of the features. 
Moreover, 
we introduce a prediction consistency regularization mechanism to assist object detection, 
which uses the multi-label prediction results as an auxiliary regularization information
to ensure consistent object category discoveries between 
the object recognition task and the object detection task.
Experiments are conducted on a few benchmark datasets 
	and the results show the proposed model outperforms
the state-of-the-art comparison methods.
\end{abstract}

\section{Introduction}
\begin{figure}[t]
\begin{center}
\includegraphics[width=4.5in,height=2.0in]{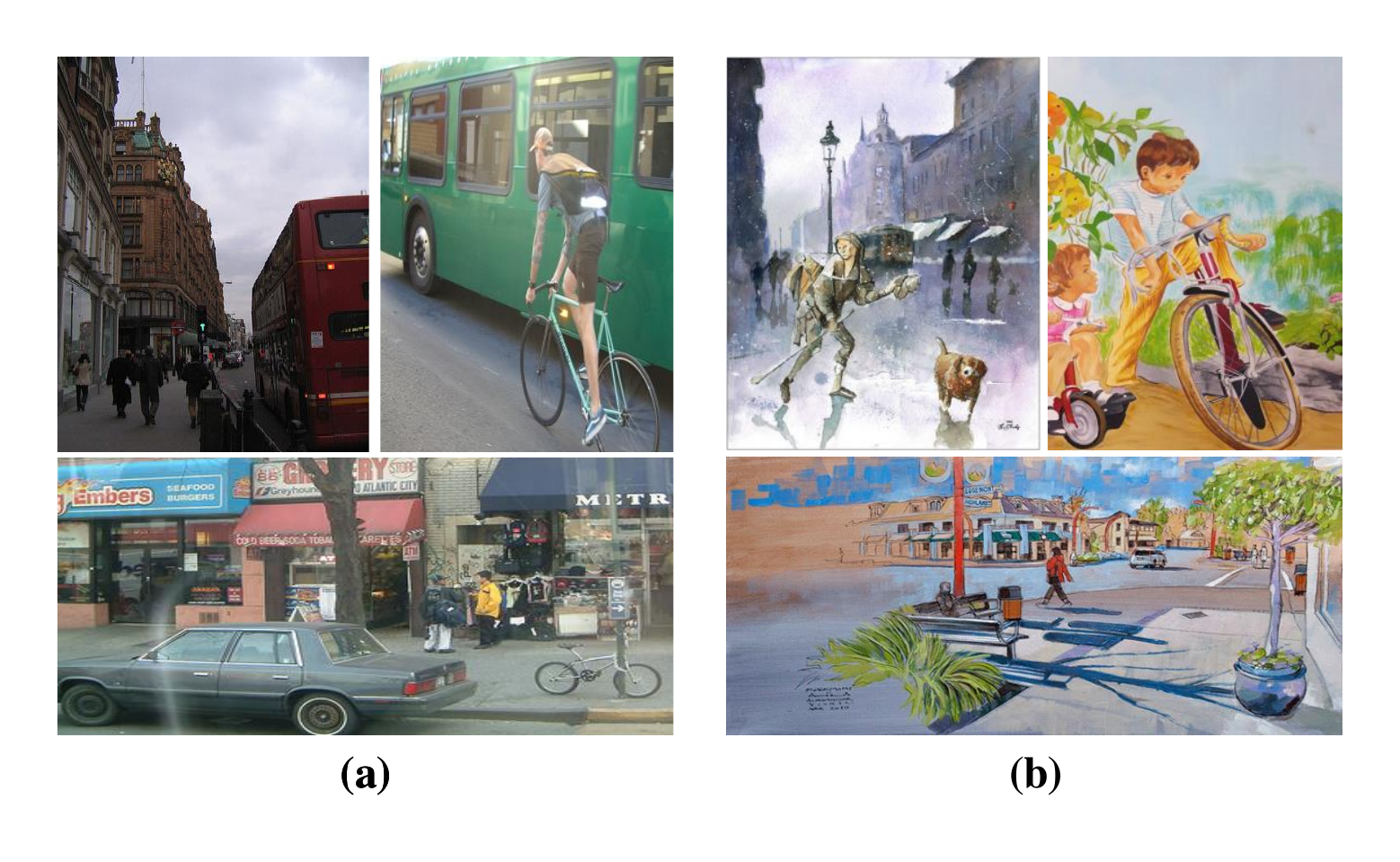}
\end{center}
   \caption{ 
   (a) and (b) are images from real scenes and virtual scenes respectively.
	It is obvious that the visual appearances of the images from different domains are very different,
	even if they contain the same categories of objects.
}
\label{fig:one}
\end{figure}

The success of deep learning models has 
led to great advances for many computer vision tasks, including
image classification~\cite{simonyan2014very,szegedy2015going,he2016deep}, 
image segmentation~\cite{Long_2015_CVPR,Zhao_2017_CVPR} 
and object detection~\cite{girshick2015fast,ren2015faster,liu2016ssd,redmon2018yolov3}. 
The smooth deployment of the deep models 
typically assumes a standard supervised learning setting,
where a sufficient amount of labeled data is available for model training
and the training and test images come from the same data source and distribution.
However, in practical applications, the training and test images can come from 
different domains that exhibit obvious deviations. 
For example, Figure~\ref{fig:one} demonstrates images 
from domains with different image styles, which obviously present 
different visual appearances and data distributions.
The violation of the i.i.d sampling principle across training and test data
prevents effective deployment of supervised learning techniques,
while acquiring new labeled data in each test domain is costly and impractical. 
To address this problem, unsupervised domain adaptation has recently received
increasing attention
\cite{ganin2016domain,tzeng2017adversarial,long2018conditional,cicek2019unsupervised}. 

Unsupervised domain adaptation aims to adapt information from a label-rich source domain 
to learn prediction models in a target domain that only has unlabeled instances. 
Although many unsupervised domain adaptation methods have been developed for simpler 
image classification and segmentation tasks
\cite{ganin2016domain,long2018conditional,cicek2019unsupervised,zhang2017curriculum,tsai2018learning,tsai2019domain}, 
much fewer domain adaptation works have been done on the more complex object detection task,
which requires recognizing both the objects and their specific locations.
The authors of \cite{chen2018domain} propose a domain adaptive faster R-CNN model for cross-domain object detection,
which employs the 
adversarial domain adaptation technique~\cite{ganin2016domain} 
to align cross-domain features at both the image-level and instance-level to bridge data distribution gaps.
This adaptive faster R-CNN method presents some promising good results.
However, due to the typical presence of multiple objects in each image, as shown in Figure~\ref{fig:one},
both the image-level and instance-level feature alignments can be problematic without considering the specific objects contained.
The more recent work~\cite{saito2019strong} proposes to address the problem of global (image-level) feature alignment 
by incorporating an additional local feature alignment under a strong-weak alignment framework for cross-domain object detection,
which effectively improved the performance of the domain adaptive faster R-CNN.
Nevertheless, this work still fails to take the latent object category information into account
for cross-domain feature alignment. 
With noisy background and various objects, 
a whole image can contain very complex information
and the overall features of an image can have complex multimodal structures. 
Aiming to learn an accurate object detector in the target domain, it is important to induce feature representations 
that minimize the cross-domain feature distribution gaps, while preserving the cross-category feature distribution gaps.

In light of the problem analysis above, 
in this paper we propose a novel end-to-end unsupervised deep domain adaptation model, 
Multi-label Conditional distribution Alignment and detection Regularization model (MCAR),
for multi-object detection,
where the images in the target domain are entirely unannotated.
The model exploits multi-label prediction as an auxiliary dual task 
to reveal the object category information in each image
and then uses this information as an additional input to perform conditional adversarial cross-domain feature alignment.
Such a conditional feature alignment is expected to improve the discriminability of the induced features
while bridging the cross-domain representation gaps to increase the transferability and domain invariance of features. 
Moreover, as object recognition is typically easier to solve and can yield higher accuracy 
than the more complex object detection task,
we introduce a consistency regularization mechanism to assist object detection, 
which uses the multi-label prediction results as auxiliary regularization information
for the object detection part to ensure consistent object category discoveries between 
the object recognition task and the object detection task.

The contribution of this work can be summarized as follows: 
(1) This is the first work that exploits multi-label prediction as an auxiliary dual task
for the multi-object detection task.
(2) We deploy a novel multi-label conditional adversarial cross-domain feature alignment methodology
to bridge domain divergence while preserving the discriminability of the features.
(3) We introduce a novel prediction consistency regularization mechanism to improve the detection accuracy.
(4) We conduct extensive experiments on multiple adaptive multi-object detection tasks 
by comparing the proposed model
with existing methods, and demonstrate effective empirical results for the proposed model.

\section{Related Work}

\noindent{\bf Object Detection.} 
Detection models have benefited from using
advanced convolutional neural networks 
as feature extractors. 
Many widely used detection methods are two-stage methods based on the region of interest (ROI)
\cite{girshick2014rich,girshick2015fast,ren2015faster}.
The RCNN in~\cite{girshick2014rich} is the first detection model 
that deploys the ROI for object detection. 
It extracts features independently from each region of interest in the image, 
instead of using the sliding window and manual feature design in traditional object detection methods. 
Later, the author of~\cite{girshick2015fast} proposed a Fast-RCNN detection model,
which adopts a ROI pooling operation to share the convolution layers between all ROIs 
and improve the detection speed and accuracy.
The work in~\cite{ren2015faster} made further improvements and proposed the Faster-RCNN, 
which combines Region Proposal Network (RPN) with Fast-RCNN to 
replace selective search and further improve detection performance.
Faster-RCNN provides a foundation for many subsequent research studies
~\cite{liu2016ssd,dai2016r,lin2017feature,he2017mask,redmon2018yolov3}.
In this work and many related unsupervised domain adaptation methods, 
the widely used two-stage method, Faster-RCNN, is adopted as the backbone detection model.
\\

\noindent{\bf Unsupervised Domain Adaptation.} 
Unsupervised domain adaptation has attracted a lot of attention in computer vision research community
and made great progress
~\cite{ganin2016domain,russo2018source,long2018conditional,kulis2011you,dziugaite2015training,shen2017wasserstein}. 
The main idea employed in these works is to learn feature representations that 
align distributions across domains.
For example, the work in~\cite{ganin2016domain} adopts the principle of  
generative adversarial networks (GANs)~\cite{goodfellow2014generative}
through a gradient reversal layer (GRL)~\cite{ganin2014unsupervised}
to achieve cross-domain feature alignment. 
The work in~\cite{long2018conditional} further extends adversarial adaptation
into conditional adversarial domain adaptation by taking the classifier's prediction into account.
The works in~\cite{russo2018source,choi2019self} use image generation to realize
cross-domain feature transformation 
and align the source and target domains. 
Moreover, some other works adopt distance metric learning methods, such as 
asymmetric metric learning~\cite{kulis2011you}, maximum mean discrepancy (MMD) minimization~\cite{dziugaite2015training} 
and Wasserstein distance minimization~\cite{shen2017wasserstein}, to achieve domain alignment. 
Nevertheless, these studies focus on the simpler image classification and segmentation tasks. 
\\

\noindent{\bf Adaptive Object Detection.} 
Recently domain adaptation for object detection has started drawing attention. 
The work in~\cite{chen2018domain} proposes an adaptive Faster-RCNN method
that uses adversarial gradient reversal to 
achieve image-level and instance-level feature alignment for adaptive cross-domain object detection.
\cite{inoue2018cross} adopts image transformation and exploits pseudo labels to realize a weakly supervised cross-domain detection. 
The work in \cite{kim2019diversify} leverages multi-style image generation between multiple domains to achieve cross-domain object detection. 
The authors of \cite{saito2019strong} propose a strong and weak alignment of local and global features 
to improve cross-domain object detection performance. 
\cite{zhu2019adapting} focuses on relevant areas for selective cross-domain alignment. 
\cite{HeMulti} adopts hierarchical domain feature alignment while adding a scale reduction module 
and a weighted gradient reversal layer to achieve domain invariance. 
\cite{cai2019exploring} advances the Mean Teacher paradigm with object relations for cross-domain detection.
\cite{shen2019scl} uses a gradient detach based multi-level feature alignment strategy for cross-domain detection. 
\cite{xie2019multi} adopts multi-level feature adversary to achieve domain adaptation. 
Nevertheless, these methods are limited to cross-domain feature alignment,
while failing to take the latent object category information into account when performing feature alignment.
Our proposed model employs multi-label object recognition as an auxiliary task and 
uses it to achieve conditional feature alignment and detection regularization.


\begin{figure*}[th!]
\begin{center}
\includegraphics[width=1.0\linewidth]{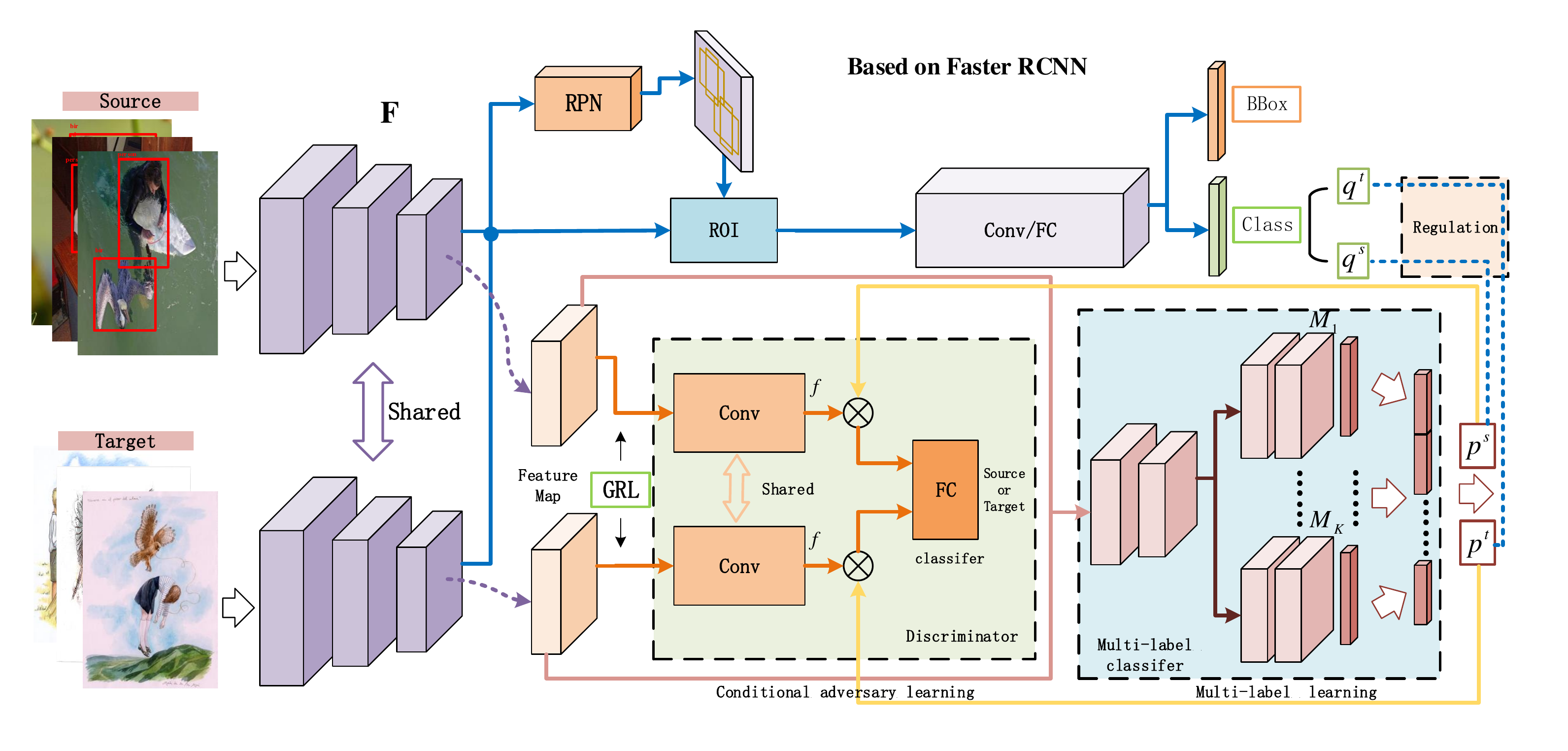}
\end{center}
   \caption{The structure of the proposed MCAR model.
  Conditional adversarial global feature alignment is conducted through a domain discriminator by
	using multi-label prediction results as object category input. 
Meanwhile, multi-label prediction results are also used to provide a prediction consistency regularization
	mechanism on object detection after the RPN.} 
\label{fig:three}
\end{figure*}

\section{Method}

In this section, we present the proposed Multi-label Conditional distribution
Alignment and detection Regularization model (MCAR) for cross-domain adaptive object detection.
We assume there are two domains from different sources and with different distributions.
The source domain is fully annotated for object detection and the target domain is entirely unannotated.
Let $X_s=\{(x_i^s,{\bf b}_i^s,{\bf c}_i^s)\}_{i=1}^{n_s}$ denote the annotated images from the source domain,
where $x_i^s$ denotes the $i$-th image, ${\bf b}_i^s$ and ${\bf c}_i^s$ denote the bounding boxes' coordinates
and the category labels of the corresponding objects contained in the image respectively.
Let $X_t=\{x_i^t\}_{i=1}^{n_t}$ denote the unannotated images from the target domain.
We assume in total $K$ classes of objects are presented in images of both the source and target domains.
We aim to train an object detection model by exploiting the available data from both domains 
such that the model can have good detection performance in the target domain.

The main idea of the proposed MCAR model is to exploit multi-label prediction (for multi-object recognition)
as an auxiliary task and use it to perform both conditional adversarial cross-domain
feature alignment and prediction consistency regularization for the target object detection task.
This end-to-end deep learning model adopts the widely used Faster-RCNN as the backbone detection network. 
Its structure is presented in Figure~\ref{fig:three}. 
Following this structure, we present the model in detail below.

\subsection{Multi-Label Prediction}\label{3.1}

The major difference between object recognition and object detection lies in that
the former task only needs to recognize the presence of any object category in the given image,
while the latter task needs to identify each specific object and its location in the image.
The cross-domain divergence in image features that impacts the object recognition
task can also consequently degrade the detection performance, 
since it will affect the region proposal network and the regional local object classification.
Therefore we propose to deploy a simpler task of object recognition to 
help extract suitable image-level features that 
can bridge the distribution gap between the source and target domains,
while being discriminative for recognizing objects. 

In particular, we treat the object recognition task as a multi-label prediction problem~\cite{zhang2006multilabel,gong2013deep}.
It takes the global image-level features produced by the feature extraction network $F$
of the Faster-RCNN model as input, and predicts the presence of $K$ object category
using $K$ binary classifier networks, $M_1, \cdots, M_K$.
These classifiers can be learned on the annotated images in the source domain,
where the global object category label indicator vector ${\bf y}_i^s\in\{0,1\}^{K}$ for the $i$-th image
can be gathered from its bounding boxes' labels ${\bf c}_i^s$ 
through a fixed transformation operation function $\varphi: {\bf c}_i^s\rightarrow{\bf y}_i^s$,
which simply finds all the existing object categories in ${\bf c}_i^s$
and represents their presence using ${\bf y}_i^s$.
The multi-label 
classifiers can then be learned by minimizing 
the following cross-entropy loss:
\begin{align}
\!\!\!
\mathcal{L}_{multi} 
&=- \frac{1}{n_s}\sum_{i=1}^{n_s}
	\left[{\bf y}_{i}^{s\top}\!\log ({\bf p}_{i}^s)
	+(1\!-\!{\bf y}_{i}^{s})^\top\! \log (1\!-\!{\bf p}_{i}^s)\right]
\end{align}
where each $k$-th entry of the prediction output vector ${\bf p}_{i}^s$
is produced from the $k$-th binary classifier:
\begin{align}
	{\bf p}_{ik}^s &= M_k(F(x_i^s))
\end{align}
which indicates the probability of the presence of objects from the $k$-th class. 

The multi-label classifiers work on the global features extracted before the RPN of the Faster-RCNN.
For Faster-RCNN based object detection, these global features will be used through RPNs
to extract region proposals and then perform object classification and bounding box regression on the proposed regions.
In the source domain, supervision information such as bounding boxes and the object labels are provided
for training the detector,
while in the target domain, the detection is 
purely based 
on the global features extracted and the detection model parameters (for RPN, region classifiers and regressors) 
obtained in the source domain.
Hence it is very important to bridge the domain gap at the global feature level.
Moreover, image features that led to good global object recognition performance
are also expected to be informative for the local object classification on proposed regions.
Therefore we will exploit multi-label prediction for global feature alignment
and regional object prediction regularization.

\subsection{Conditional Adversarial Feature Alignment}\label{3.2}

The popular generative adversarial network (GAN)~\cite{goodfellow2014generative} has shown that 
two distributions can be aligned by using a discriminator as an adversary to play a minimax two-player game. 
Following the same principle, conditional adversary is designed to 
take label category information into account. 
It has been suggested in~\cite{long2018conditional,mirza2014conditional} 
that the cross-covariance of the predicted category information and the global image features 
can be helpful for avoiding partial alignment and achieving multimodal feature distribution alignment.
We propose to integrate the multi-label prediction results
together with the global image features extracted by $F$ to perform conditional adversarial
feature alignment at the global image level.
The key component network introduced is the domain discriminator $D$,
which predicts the domain of the input image instance,
with label 1 indicating the source domain and 0 indicating the target domain.
As shown in Figure~\ref{fig:three}, the discriminator 
consists of a convolution filter layer $f$, which reduces the dimension of the input features, 
and a fully connected layer $FC$, which integrates the inputs to perform classification.
It takes features $F(x_i)$ and the multi-label prediction probability vector ${\bf p}_i$ as input,
and uses softmax activation function to produce probabilistic prediction output. 
For the conditional adversarial training, we adopted a focal loss
~\cite{lin2017focal,saito2019strong}, which uses the prediction confidence deficiency score to weight 
each instance in order to give more weights to hard-to-classify examples. 
The loss of conditional adversarial training, $\mathcal{L}_{adv}$, is as below:
\begin{align}
	\label{eq:adv}	
	\min_{F} &\max_{D} \quad \mathcal{L}_{adv} = - \frac{1}{2}(\mathcal{L}_{adv}^s+\mathcal{L}_{adv}^t)
\\	
\mathcal{L}_{adv}^s &\!=\!  - \frac{1}{n_s}\!\sum\limits_{i = 1}^{n_s} 
	{{{(1 \!-\! D(F(x_i^s),{\bf p}_i^s))}^\gamma }
	\log (D(F(x_i^s), {\bf p}_i^s))}
\nonumber \\	
\mathcal{L}_{adv}^t &\!=\!  - \frac{1}{n_t}\!\sum\limits_{i = 1}^{n_t} 
	{D{{(F(x_i^t),{\bf p}_i^t)}^\gamma }
	\log(1 \!-\! D(F(x_i^t),{\bf p}_i^t))}
\nonumber	
\end{align}
where $\gamma$ is a modulation factor that controls how much to focus on the hard-to-classify example;
the global features $F(x_i)$ and the multi-label prediction probability vector ${\bf p}_i$ are
integrated through a multi-linear mapping function such that
$D(F(x_i), {\bf p}_i) = FC(f(F(x_i))\otimes  {\bf p}_i)$. 
With this adversary loss, 
the feature extractor $F$ will be adjusted to try to confuse the domain 
discriminator $D$, while $D$ aims to maximumly separate the two domains. 

This multi-label prediction conditioned adversarial feature alignment 
is expected to bridge the domain distribution gaps while preserving the discriminability for object recognition,
which will improve 
the adaptation of 
the consequent region proposal, 
object classification on each proposed region and its location identification in the target domain. 

\subsection{Category Prediction based Regularization}
\label{3.3}

The detection task involves recognizing both the objects and their locations,
which is relatively more difficult than object recognition
\cite{everingham2010pascal}.
The multi-label classifiers we applied can produce more accurate recognition results
as the region proposal mistakes can be accumulated to objection classification 
on the proposed regions in the detection task.
Based on such an observation, we propose a 
novel category prediction consistency regularization
mechanism for object detection by exploiting multi-label prediction results.

Assume $N$ region proposals are generated through the region proposal network (RPN) for an input image $x$.
Each proposal will be classified into one of the $K$ object classes using an object classifier $C$,
while its location coordinates will be produced using a regressor $R$. 
The multi-class object classifier produces a length $K$ prediction vector $\hat{\bf q}$ 
on each proposal
that indicates the probability of the proposed region belonging to one of the $K$ object classes.
The object prediction on the total $N$ proposals can form a prediction matrix $Q\in[0,1]^{K\times N}$.
We can then compute an overall multi-object prediction probability vector ${\bf q}$ by taking the row-wise maximum over $Q$,
such that ${\bf q}_k = \max(Q(k,:))$, and use 
${\bf q}_k$ as the prediction probability of the image $x$ containing the $k$-th object category.
To enforce consistency between the prediction produced by the detector and the prediction
produced by the multi-label object recognition,
we propose to minimize the $KL$ divergence between their prediction probability vectors ${\bf p}$ and ${\bf q}$
after renormalizing each vector with softmax function.
As $KL$ divergence is an asymmetric measure, 
we define the consistency regularization loss as:
\begin{align}
\mathcal{L}_{kl} &= \mathcal{L}^s_{kl} + \mathcal{L}^t_{kl}	
\\	
\mathcal{L}^s_{kl}&= \frac{1}{2n_s}\sum_{i=1}^{n_s} (KL({\bf p}^s_i,{\bf q}^s_i)+KL({\bf q}^s_i,{\bf p}^s_i))  
\\
\mathcal{L}^t_{kl}&= \frac{1}{2n_t}\sum_{i=1}^{n_t} (KL({\bf p}^t_i,{\bf q}^t_i)+KL({\bf q}^t_i,{\bf p}^t_i))  
\end{align}
With this regularization loss, we expect the multi-label prediction results
can assist object detection through unified mutual learning.

\subsection{Overall End-to-End Learning}

The detection loss of the base Faster-RCNN model, denoted as $\mathcal{L}_{det}$,
is computed on the annotated source domain data under supervised classification and regression.
It has two components, the proposal classification loss and the bounding box regression loss. 
We combine the detection loss, the multi-label prediction loss, 
the conditional adversarial feature alignment loss, and the prediction consistency regularization loss 
together for end-to-end deep learning. The total loss can be written as:
\begin{equation}
\begin{split}
	\label{eq:loss}	
\left\{ {\begin{array}{*{20}{c}}
{{\mathcal{L}_{all}} = 
{\mathcal{L}_{\det }} + \lambda {\mathcal{L}_{adv}} + \mu {\mathcal{L}_{multi}} + \varepsilon {\mathcal{L}_{kl}}}
	\\\\
	{\mathop {\min }\limits_F \mathop {\max }\limits_D\quad {\mathcal{L}_{all}}}
\end{array}} \right.
\end{split}
\end{equation}
where $\lambda$, $\mu$, and $\varepsilon$ 
are trade-off parameters that balance the multiple loss terms.
We use SGD optimization algorithm to perform training,
while GRL~\cite{ganin2014unsupervised} is adopted to implement the gradient sign flip for
the domain discriminator part.



\begin{table*}[t]
\begin{center}
\caption{Test results of domain adaptation for object detection from PASCAL VOC to Watercolor in terms of 
	mean average precision (\%). 
	MC and PR indicate 
	Multilabel-Conditional adversary 
	and Prediction based Regularization, respectively.
	}
\renewcommand\arraystretch{1.2}
\setlength{\tabcolsep}{4pt}
{
\begin{tabular}{l|cc|cccccc|c}
\hline
	Method                & MC & PR  & bike & bird & car  & cat  & dog  & person & mAP  \\ \hline
Source-only           &     &    & 68.8 & 46.8 & 37.2 & 32.7 & 21.3 & 60.7   & 44.6 \\ \hline
BDC-Faster~\cite{saito2019strong}              &     &    & 68.6 & 48.3 & 47.2 & 26.5 & 21.7 & 60.5   & 45.5 \\ \hline
DA-Faster~\cite{chen2018domain}              &     &    & 75.2 & 40.6 & 48.0 & 31.5 & 20.6 & 60.0   & 46.0 \\ \hline
SW-DA~\cite{saito2019strong}               &  &    & 82.3 & \bf 55.9 & 46.5 & 32.7 &  35.5 & 66.7   & 53.3 \\ \hline
SCL~\cite{shen2019scl} &   &    
	& 82.2 & 55.1 & {\bf 51.8} & 39.6 & \bf38.4 & 64.0   & 55.2 \\ \hline
	\multirow{2}{*}{MCAR (Ours)}   & \checkmark  &    & \bf 92.5 & 52.2 & 43.9 & \bf 46.5 & 28.8 & 62.5   & 54.4 \\ \cline{2-10} 
                          & \checkmark   & \checkmark & 87.9 & 52.1 & \bf 51.8 & 41.6 & 33.8 & \bf 68.8   & \bf 56.0 \\ \hline\hline
Train-on-Target                 &     &    & 83.6 & 59.4 & 50.7 & 43.7 & 39.5 & 74.5   & 58.6 \\ \hline
\end{tabular}}
\label{tab:one}
\end{center}
\end{table*}

\begin{table}[t]
\begin{center}
\caption{Test results of domain adaptation for object detection from PASCAL VOC to Comic, 
	The definition of MC and PR is same as in Table~\ref{tab:one}.}
\renewcommand\arraystretch{1.2}
\setlength{\tabcolsep}{4pt}
{
\begin{tabular}{l|cc|cccccc|c}
\hline
Method                & MC & PR  & bike & bird & car  & cat  & dog  & person & mAP  \\ \hline
Source-only           &    &    & 32.5    & 12.0    & 21.1    & 10.4    & 12.4    & 29.9      & 19.7    \\ \hline
DA-Faster         &    &    & 31.1 & 10.3 & 15.5 & 12.4 & 19.3 & 39.0   & 21.2 \\ \hline
SW-DA           &    &    & 36.4 & 21.8 & 29.8 & 15.1  & 23.5 & 49.6   & 29.4 \\ \hline
	\multirow{2}{*}{MCAR (Ours)} & \checkmark &    & 40.9 & \bf22.5 & 30.3 & \bf23.7 & \bf24.7 & \bf53.6   & 32.6 \\ \cline{2-10} 
                      & \checkmark & \checkmark & \bf47.9 & 20.5 & \bf37.4 & 20.6 & 24.5 & 50.2   & \bf33.5 \\ 

\hline
\end{tabular}}
\label{tab:two}
\end{center}
\end{table}

\section{Experiments}
We conducted experiments with multiple cross-domain multi-object detection tasks under different adaptation scenarios: 
(1) Domain adaptation from real to virtual image scenarios, where we used cross-domain detection tasks 
from PASCAL VOC~\cite{everingham2010pascal} to Watercolor2K~\cite{inoue2018cross} and Comic2K~\cite{inoue2018cross} respectively.
(2) Domain adaption from normal/clear images to foggy image scenarios, where
we used object detection tasks that adapt from 
Cityscapes~\cite{cordts2016cityscapes} to Foggy Cityscapes~\cite{sakaridis2018semantic}. 
In each adaptive object detection task, the images in the source domain are fully annotated and 
the images in the target domain are entirely unannotated. 
We present our experimental results and discussions in this section.

\subsection{Implementation Details}

In the experiments, we followed the setting of ~\cite{saito2019strong} 
by using the Faster-RCNN as the backbone detection network, 
pretraining the model weights on the ImageNet,
and using the same 600 pixels of images' shortest side. 
We set the training epoch as 25, 
and set $\lambda$, $\mu$, $\varepsilon$, and  $\gamma$ as 0.5, 0.01, 0.1, and 5 respectively. 
The momentum is set as 0.9 and weight decay as 0.0005. 
For all experiments, we evaluated different methods using mean average precision (mAP) with a threshold of 0.5. 
By default, in the multi-label learning, 
all the convolutional layers have 3x3 convolution kernels and 512 channels.
The convolution layer in conditional adversary learning also has 3x3 convolution kernel and 512 channels. 
These convolution parameters can be adjusted to suit different tasks, but our experiments all adopt the default setting,
which yield good results.

\begin{table*}[t]
\begin{center}
	\caption{Test results of domain adaptation for object detection from Cityscapes to Foggy Cityscapes in terms of mAP (\%). 
	MC and PR are same as in Table~\ref{tab:one}.}
\renewcommand\arraystretch{1.2}
{
\begin{tabular}{l|cc|cccccccc|c}
\hline
Method                & MC                    & PR  & person & rider & car  & truck & bus  & train & motorbike & bicycle & mAP  \\ \hline
Source-only           &                       &    & 25.1   & 32.7  & 31.0 & 12.5  & 23.9 & 9.1   & 23.7      & 29.1    & 23.4 \\ \hline
BDC-Faster~\cite{saito2019strong}            &                       &    & 26.4   & 37.2  & 42.4 & 21.2  & 29.2 & 12.3  & 22.6      & 28.9    & 27.5 \\ \hline
DA-Faster~\cite{chen2018domain}              &                       &    & 25.0   & 31.0  & 40.5 & 22.1  & 35.3 & 20.2  & 20.0      & 27.1    & 27.6 \\ \hline
SC-DA~\cite{zhu2019adapting}           &                       &    & 33.5   & 38.0  & \bf 48.5 & 26.5  & 39.0 & 23.3  & 28.0      & 33.6    & 33.8 \\ \hline
MAF~\cite{HeMulti} &    & & 28.2   & 39.5  & 43.9  & 23.8  & 39.9  & 33.3   & 29.2       & 33.9    & 34.0 \\ \hline
SW-DA~\cite{saito2019strong}             &                       &    & \bf 36.2   & 35.3  & 43.5 & 30.0  & 29.9 & 42.3  & 32.6      & 24.5    & 34.3 \\ \hline
DD-MRL~\cite{kim2019diversify} &    &&30.8 &40.5 &44.3 &27.2 & 38.4 &34.5 & 28.4 &32.2 &34.6 \\ \hline
MTOR~\cite{cai2019exploring} &    &&30.6 &41.4 &44.0 &21.9 & 38.6 &40.6 & 28.3  & 35.6  & 35.1 \\ \hline
Dense-DA~\cite{xie2019multi}             &                       &    & 33.2   & \bf 44.2  & 44.8 & 28.2  & 41.8 & 28.7  & 30.5      & 36.5    & 36.0 \\ \hline
SCL~\cite{shen2019scl}&    & & 31.6 &44.0 &44.8 &30.4 &41.8 &40.7 &33.6 &36.2 &37.9 \\ \hline
	\multirow{2}{*}{MCAR (Ours)} & \checkmark &    & 31.2   & 42.5  & 43.8     & \bf 32.3  &  41.1    &   33.0    &  32.4    &36.5  & 36.6  \\ \cline{2-12} 
                      & \checkmark& \checkmark & 32.0     & 42.1   & 43.9     & 31.3    & \bf 44.1  & \bf 43.4  & \bf 37.4  & \bf36.6    & \bf 38.8     \\ \hline\hline
Train-on-Target      & \multicolumn{1}{c}{} &    & 50.0      & 36.2     & 49.7    & 34.7    & 33.2    & 45.9    & 37.4    &  35.6   & 40.3    \\ \hline
\end{tabular}}
\label{tab:three}
\end{center}
\end{table*}


\subsection{Domain Adaptation from Real to Virtual Scenes}

In this set of experiments, 
we used the PASCAL VOC~\cite{everingham2010pascal} dataset as the source domain, 
and used the Watercolor2k and Comic2k~\cite{inoue2018cross} as the target domains. 
PASCAL VOC contains realistic images, while Watercolor2k and Comic2k contain virtual scene images. 
There are significant differences between the source and target domains. 
The training set of PASCAL VOC (Trainval of PASCAL VOC 2007 and PASCAL VOC 2012) includes 
20 different object labels and a total of 16,551 images. 
Watercolor2k and Comic2k contain 6 different classes (`bicycle', `bird', `car', `cat', `Dog', `person'), each providing 2K images, and splitting equally into training and test sets. 
These 6 categories are included in the 20 categories of PASCAL VOC. 
We used the 1K training set in each target domain 
for training the domain adaptation model,
while evaluating the model and report results with the 1K test set.
In this experiment, 
we used resnet101~\cite{he2016deep} as the backbone network of the detection model.
\\

\noindent{\bf PASCAL VOC to Watercolor.} 
The test detection results yield by 
adaptation from PASCAL VOC to Watercolor are reported in Table~\ref{tab:one}. 
Our proposed MCAR model is compared with 
the source-only baseline and the state-of-the-art adaptive object detection methods, including
BDC-Faster~\cite{saito2019strong},
DA-Faster~\cite{chen2018domain}, 
SW-DA~\cite{saito2019strong},
and SCL~\cite{shen2019scl}. 
The Train-on-Target results, obtained by training on labeled data in the target domain, are provided
as upperbound reference values.
We can see under the same experimental conditions, 
our proposed method achieves the best overall result, 
while only underpeforming the Train-on-Target by 2.6\%. 
Comparing to source only, our method achieves a remarkable overall performance improvement of 9.8\%. 
Although SW-DA~\cite{saito2019strong} confirmed the validity of local and global feature alignment and showed a significant performance improvement over other methods, our method surpasses SW-DA by 2.7\%. 
Meanwhile, our method also outperforms SCL~\cite{shen2019scl} which relies on stacked multi-level feature alignment.
The results suggest the proposed multi-label learning based feature alignment and prediction regularization
are effective.
\\

\noindent{\bf PASCAL VOC to Comic.} 
The results of adaptation from PASCAL VOC to Comic 
are reported in Table~\ref{tab:two}. 
Again, the proposed MCAR method achieved the best adaptive detection result.
It outperforms the baseline, source-only (trained on source domain data without any adaptation), by 13.8{\%}, 
and outperforms the best comparison method, SW-DA, by 4.1\%,
These results again show that our model is very suitable for adaptive multi-object detection.

\subsection{Adaptation from Clear to Foggy Scenes.}
In this experiment, we perform adaptive object detection from normal clear images to foggy images.
We use the Cityscapes dataset as the source domain. Its images came from 27 different urban scenes, 
where the annotated bounding boxes are generated by the original pixel annotations.
We use the Foggy Cityscapes dataset as the target domain. 
Its images have been rendered by Cityscapes, which can simulate fog in real road conditions with deep rendering. 
They contain 8 categories: `person', `rider', `car', `truck', `bus', `train', `motorcycle' and `bicycle'. 
In this experiment, we used vgg16~\cite{simonyan2014very} as the backbone of the detection model.
We recorded the test results on the validation set of Foggy Cityscapes.

The results are reported in the Table~\ref{tab:three}. 
We can see the proposed MCAR method achieved the best adaptive detection result.
It outperforms source-only by 15.4{\%}, and outperforms the two best comparison methods, 
Dense-DA~\cite{xie2019multi} and SCL~\cite{shen2019scl}, 
by 2.8\% and 0.9\%.
Moreover, it is worth noting that the performance of the proposed approach is very close to the Train-on-Target; 
the result of the Train-on-Target is only 1.5\% higher than ours.
Due to the very complex road conditions in this task, 
although the multi-label classifier is more capable of category judgment than the detection model, 
its accuracy is not much higher.
Hence in this experiment, we used the combination of the multi-label category prediction
and the object detection level category prediction. 
That is, we used $softmax({\bf p}+{\bf q})$ as the label category information
for the conditional adversarial feature alignment.
This experiment presents and validates a natural variant of the proposed model.

\begin{table*}[t]
\begin{center}
	\caption{The ablation study results in terms of mAP(\%) 
	on the adaptive detection task of Cityscapes $\rightarrow$ Foggy Cityscapes. 
``w/o-adv" indicates dropping the conditional adversary loss; 
``uadv" indicates replacing the conditional adversary loss with an unconditional adversary loss;
``w/o-PR" indicates dropping the prediction regularization loss;
and 
``w/o-MP-PR" indicates dropping both the multilabel prediction loss and the prediction regularization loss.
}
\label{tab:six}
\renewcommand\arraystretch{1.2}
{
\begin{tabular}{l|cccccccc|c}
\hline
Method   & person & rider & car  & truck & bus  & train & motorbike & bicycle & mAP  \\ \hline
MCAR
& 32.0   & 42.1  & 43.9 & 31.3  & 44.1 & 43.4   & 37.4      & 36.6    & \bf38.8\\\hline
MCAR-w/o-PR
& 31.2   & 42.5  & 43.8 & 32.3  & 41.1 & 33.0   & 32.4      & 36.5    & 36.6\\\hline
MCAR-uadv
& 31.7   & 42.0  & 45.7 & 30.4  & 39.7 & 14.9   & 28.6     & 36.5    & 33.7\\\hline
MCAR-uadv-w/o-PR
& 32.8   & 40.1  & 43.8 & 23.0  & 30.9 & 14.3   & 30.3      & 33.1    & 31.0\\\hline
MCAR-uadv-w/o-MP-PR
& 30.5   & 43.2  & 41.4 & 21.7  & 31.4 & 13.7   & 29.8      & 32.6    & 30.5\\\hline
MCAR-w/o-adv& 25.0   & 34.9  & 34.2 & 13.9  & 29.9 & 10.0   & 22.5      & 30.2    & 25.1\\

\hline

\end{tabular}}
\end{center}
\end{table*}

\subsection{Ablation Study}

The proposed MCAR model has two major mechanisms, Multilabel-conditional adversary (MC) and Prediction based Regularization (PR), which are incorporated into the learning process through 
the three auxiliary loss terms in Eq.(\ref{eq:loss}): 
the conditional adversary loss $\mathcal{L}_{adv}$, 
the multi-label prediction loss $\mathcal{L}_{multi}$, 
and the prediction regularization loss $\mathcal{L}_{kl}$. 
The conditional adversary loss uses the multi-label prediction outputs as its conditions,
and hence the two loss terms, $\mathcal{L}_{adv}$ and $\mathcal{L}_{multi}$,
together form the multilabel-conditional adversary (MC),
while the prediction regularization (PR) is also built on the multi-label prediction outputs
through the regularization loss $\mathcal{L}_{kl}$. 
To investigate the impact of these loss components, 
we conducted a more comprehensive ablation study 
on the adaptive detection task from Cityscapes to Foggy Cityscapes
by comparing MCAR with its multiple variants. 
The variant methods and results are reported in Table~\ref{tab:six}.

We can see that dropping the conditional adversary loss ({\em MCAR-w/o-adv}) 
leads to large performance degradation. 
This makes sense since the adversarial loss
is the foundation for cross-domain feature alignment. 
By replacing the conditional adversary loss with an unconditional adversary loss,
{\em MCAR-uadv} loses the multilabel-conditional adversary (MC) component, 
which leads to remarkable performance degradation 
and verifies the usefulness of 
the multi-label prediction based cross-domain multi-modal 
feature alignment. 
Dropping the prediction regularization loss 
from either {\em MCAR}, which leads to {\em MCAR-w/o-PR}, 
or {\em MCAR-uadv}, which leads to {\em MCAR-uadv-w/o-PR},
induces additional performance degradation.
This verifies the effectiveness of the prediction regularization strategy,
which is built on the multi-label prediction outputs as well.
Moreover, 
by further dropping the multi-label prediction loss from {\em MCAR-uadv-w/o-PR},
the variant {\em MCAR-uadv-w/o-MP-PR}'s performance also drops slightly.
Overall these results validated the effectiveness of the proposed MC and PR mechanisms,
as well as the multiple auxiliary loss terms in the proposed learning objective.


\begin{figure}[ht]
\begin{center}
\includegraphics[width=0.68\linewidth]{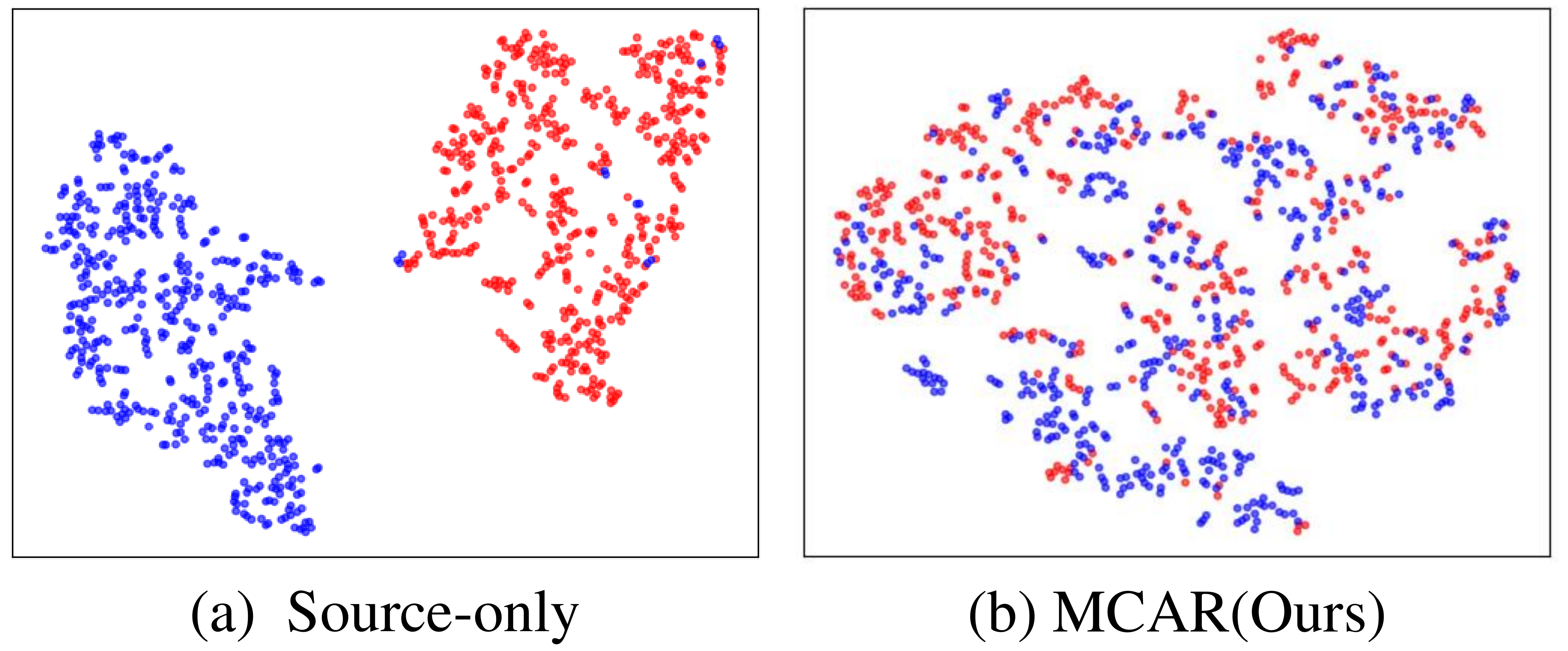}
\end{center}
\caption{Feature visualization results. (a) and (b) respectively represent the feature distribution results of the Source-only model and our model in the clear (Cityscapes) and foggy (Foggy Cityscapes) scenes. Red indicates from the source domain and blue indicates from the target domain}
\label{fig:four}
\end{figure}
\begin{figure*}[http]
\begin{center}
\includegraphics[width=1.0\linewidth]{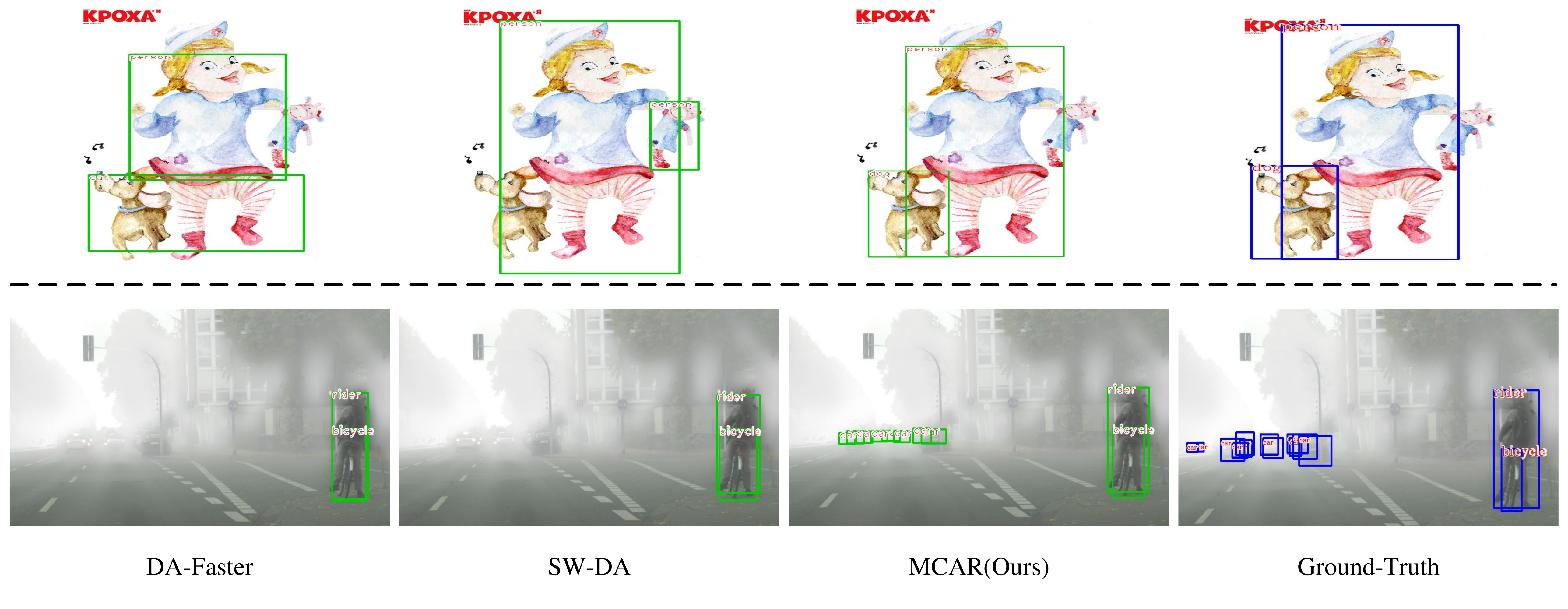}
\end{center}
   \caption{Qualitative results on adaptive detection. 
	The top row presents examples of domain adaptive detection from PASCAL VOC to Watercolor. 
	The bottom row shows examples of adaptive detection from Cityscapes to Foggy Cityscapes. 
	The green box represents the results obtained by the detection models, and the blue box represents the ground-truth annotation.}
\label{fig:five}
\end{figure*}

\begin{table}[t]
\begin{center}
\caption{Parameter sensitivity analysis on the adaptation task
	from PASCAL VOC to watercolor. }
\setlength{\tabcolsep}{4pt}
{
\begin{tabular}{c|ccccc||c|ccccc}
\hline
	$\lambda$ &\multicolumn{5}{c||}{0.5}   
	& $\gamma$ &\multicolumn{5}{c}{5}\\ \hline 
	$\gamma$ & 1 & 3 & 5 & 7 & 9
	&$\lambda$ & 0.1 & 0.25 & 0.5 & 0.75 & 1\\ \hline
	mAP & 44.0 & 46.1 & \bf54.4 & 49.1 &44.8
	& mAP & 49.1 & 50.2 & \bf54.4 & 50.1 &49.3\\ 
\hline
\end{tabular}}
\label{tab:five}
\end{center}
\end{table}


\subsection{Further Analysis}

\noindent{\bf Feature visualization.} 
On the task of adaptation from Cityscapes to Foggy Cityscapes, 
we used t-SNE~\cite{maaten2008visualizing} to compare the distribution of induced features between our model and the Source-only model (clear to fogg scenes).
The results are shown in Figure~\ref{fig:four}. We can see that with the feature distribution obtained by source-only (Figure~\ref{fig:four}(a)),  
the source domain and target domain are obviously separated, which shows the existence of domain divergence. 
By contrast, our proposed method produced features that can well confuse the domain discriminators.
This suggests that our proposed model has the capacity to bridge the domain distribution divergence and induce domain invariant features.
\\

\noindent
{\bf Parameters sensitivity analysis.} 
We conducted sensitivity analysis on the two hyperparameters, $\lambda$ and $\gamma$
using the adaption task from PASCAL VOC to Watercolor. 
$\lambda$ controls the weight of adversarial feature alignment,
while $\gamma$ controls the degree of focusing on hard-to-classify examples.
Other hyperparameters are set to their default values. 
We conducted the experiment by fixing the value of $\gamma$ to adjust $\lambda$, 
and then fixing $\lambda$ to adjust $\gamma$.
Table~\ref{tab:five} presents the results.
We can see
with the decrease of parameter $\gamma$ from its default value 5, 
the test performance degrades as
the influence of domain classifier on difficult samples 
is weakened and the contribution of easy samples is increased.  
When $\gamma=1$, it leads to the same result as the basic model,
suggesting the domain regulation ability basically fails to play its role. 
On the other hand, a very large $\gamma$ value is not good either, 
as the most difficult samples will dominate.
For $\lambda$, we find that $\lambda=0.5$ leads to the best performance. 
As detection is still the main task, it makes sense to have the $\lambda <1$. 
When $\lambda=0$, it degrades to a basic model without feature alignment. 
Therefore, some value in the middle would be a proper choice.
\\

\noindent{\bf Qualitative results.} 
Object detection results are suitable to be qualitatively judged through visualization. 
Hence we present some qualitative adaptive detection results in the target domain in Figure~\ref{fig:five}.
The top row of  Figure~\ref{fig:five} presents the qualitative detection result of three 
state-of-the-art adaptive detection methods, 
DA-Faster, SW-DA, and MCAR (ours), and the ground-truth on an image from Watercolor.
We can see both 
`DA-Faster' and `SW-DA' have some false positives, while failing to detect the object of `dog'. 
Our model correctly detected both the `person' and the `dog'.
The bottom row of Figure~\ref{fig:five} presents the detection results of the DA methods 
and the ground-truth on an image from Foggy Cityscapes. 
We can see it is obvious that the cars in the distance are very blurred and difficult to detect due to the fog.
The DA-Faster and SW-DA fail to find these cars,
while our model successfully detected them.
\section{Conclusion}

In this paper, we propose an unsupervised multi-object cross-domain detection method. 
We exploit multi-label object recognition as a dual auxiliary task
to reveal the category information of images from the global features. 
The cross-domain feature alignment is conducted by 
performing conditional adversarial distribution alignment with
the combination input of global features and multi-label prediction outputs.
We also use the idea of mutual learning 
to improve the detection performance 
by enforcing consistent object category predictions
between the multi-label prediction over global features 
and the object classification over detection region proposals.
We conducted experiments on multiple cross-domain multi-objective detection datasets.
The results show the proposed model achieved the state-of-the-art performance.

\clearpage

\bibliographystyle{splncs04}
\bibliography{egbib}
\end{document}